%% file: acl_latex.tex
\pdfoutput=1

\documentclass[11pt]{article}

\usepackage[final]{acl}

\usepackage{times}
\usepackage{latexsym}

\usepackage[T1]{fontenc}

\usepackage[utf8]{inputenc}

\usepackage{microtype}

\usepackage{inconsolata}

\usepackage{graphicx}
\usepackage{booktabs} 
\usepackage{tcolorbox}
\usepackage{makecell}

\input{commands}

\title{Coling-UniA at SciVQA 2025: Few-Shot Example Retrieval and Confidence-Informed Ensembling for Multimodal Large Language Models}

\author{
  Christian Jaumann \hspace{9mm} Annemarie Friedrich \hspace{9mm} Rainer Lienhart\\
  University of Augsburg, Germany \\
  \texttt{\{firstname.lastname\}@uni-a.de}
}

\begin{document}
\maketitle
\begin{abstract}
This paper describes our system for the SciVQA 2025 Shared Task on Scientific Visual Question Answering.
Our system employs an ensemble of two Multimodal Large Language Models and various few-shot example retrieval strategies. The model and few-shot setting are selected based on the figure and question type. We also select answers based on the models' confidence levels.
On the blind test data, our system ranks third out of seven with an average F1 score of 85.12 across ROUGE-1, ROUGE-L, and BERTS. 
Our code is publicly available.\footnote{\href{https://github.com/coling-unia/few-shot-scivqa2025}{https://github.com/coling-unia/few-shot-scivqa2025}}
\end{abstract}

\section{Introduction}
Visual Question Answering (VQA) requires systems to answer natural language questions about visual content. 
The complexity of these questions can range from binary questions to free-form and open-ended questions.
Existing VQA datasets address various types of images, e.g.,
VQA v2 focuses on real-world photos \citep{Goyal2017Making}, DocVQA focuses on scanned documents \citep{Mathew2021DocVQA}, while ChartQA \citep{masry-etal-2022-chartqa} and PlotQA \citep{Methani2020PlotQA} focus on charts.

In this paper, we describe our system submission for the 2025 Shared Task on Scientific Visual Question Answering (SciVQA) \citep{borisova-scivqa-2025}. The dataset\footnote{\href{https://huggingface.co/datasets/katebor/SciVQA}{https://huggingface.co/datasets/katebor/SciVQA}} comprises 3000 real-world scientific figure images, which were collected from the ACL-Fig \citep{Karishma2023ACL-Fig} and SciGraphQA \citep{Li2023SciGraphQA}.

Most existing VQA approaches that focus on charts rely on models explicitly tuned for this domain \citep{liu-etal-2023-matcha, Han2023ChartLlama, Xia2024ChartXChartVLM, Zhang2024TinyChart}. 
In contrast, our approach uses Multimodal Large Language Models (MLLMs) in a zero/few-shot setting without any fine-tuning.
We test several strategies for retrieving few-shot examples from the training set
based on question or question-and-image similarity. 
We find that performance varies widely by question/figure type and by MLLM.
Our best-performing approach first selects 
highly confident answers 
from a configuration of an MLLM and a few-shot setting. 
For all remaining instances, the system configuration is varied by the instance's question type.
In the official evaluation, our system ranks third.

\begin{figure}
    \centering
    \includegraphics[width=1\columnwidth]{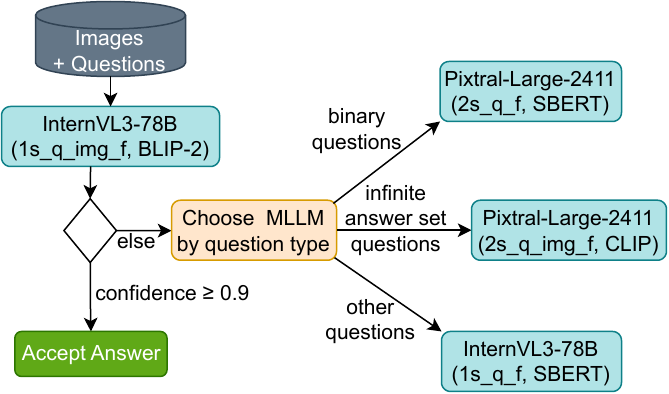}
    \caption{System overview. Abbreviations for few-shot example selection: \#s = \#-shot, q = question similarity, img = image similarity, f = filter for same figure type, nf = no filtering (search in entire train set).}
    \label{fig:system_overview}
\end{figure}

\section{Method}
Our system is configurable to use different MLLMs in either a zero-shot or a few-shot setting. 
These settings are combined using an ensemble approach (see \autoref{fig:system_overview}) that first selects all high-confidence answers from a configuration that we find to be well-calibrated, i.e., the predicted confidence scores align well with the actual empirical accuracy on the development set.
We approximate answer confidence by exponentiating the mean log-probability of all generated answer tokens.
For the remaining instances, the model configuration is selected based on question type as identified on the development set.
The MLLM is prompted with each image and the associated question (see Appendix \ref{ssec:0s_prompt}).
Following the oracle-style setup of the Shared Task, we also provide the model with additional image metadata that is included in the dataset, i.e., the image caption, figure type, and whether the image contains multiple subfigures.
The task description depends on whether there are pre-defined answer options for the questions. 
The model is instructed to answer and to determine whether it is possible to answer based solely on the provided information.




\begin{table}[t]
\footnotesize
\centering
\setlength\tabcolsep{3pt}
\begin{tabular}{llrrrr}
\toprule
\textbf{Rank} & \textbf{Submission} & \textbf{R1-F1} & \textbf{RL-F1} & \textbf{BS-F1} & \textbf{Avg.}\\
\midrule
1.& ExpertNeurons & 80.49 & 80.43 & 98.49 & 86.47 \\
2. & THAii\_LAB   &  78.99 & 78.92 & 98.39 & 85.43 \\
3. & \textbf{Coling-UniA }  & 78.62 & 78.56 & 98.17 & 85.12 \\
\hline
& Median & 75.83 & 75.75 & 98.36 & 83.31 \\
\bottomrule
\end{tabular}
  \caption{Overview of SciVQA@SDP 2025 results. Metrics: R1 = ROUGE-1, RL = ROUGE-L, BS = BERTS.}
  \label{tab:leaderboard_scores}
\end{table}

To enhance the reproducibility, our selection of MLLMs is constrained to open-weights models. We use InternVL3-78B \citep{zhu2025internvl3} and Pixtral-Large-Instruct-2411.\footnote{\href{https://huggingface.co/mistralai/Pixtral-Large-Instruct-2411}{https://huggingface.co/mistralai/Pixtral-Large-Instruct-2411}} 
We run all models using 16-bit quantization and a temperature of 0.

\paragraph{Few-shot Example Retrieval.}
We evaluate different few-shot retrieval approaches.
First, we use question similarity to select examples from the training data for the input instance.
For ranking, we use the cosine similarities of the questions' SBERT embeddings \citep{reimers-2019-sentence-bert}. 
Second, we select examples based on the question-image similarity using CLIP \citep{Radford2021Learning}. 
We compute CLIP embeddings for each question and image, normalize them,
compute the mean embedding of each image-question pair, normalize again, and determine the best-fit example using cosine similarity. 
We also experimented with computing similarities based on the image-question embeddings directly provided by BLIP-2 \citep{Li2023BLIP-2}.
In case of similarity ties, we choose the first instance in the order as they are provided in the training set.


For both settings, we retrieve few-shot examples from the training set in two variants:
(1) We consider only the
subset of the training data that has the same figure type, and, if possible, the same number of sub-figures, as the input instance.
(2) We search for few-shot examples in the entire training set. 
In both cases, we exclude all instances that use the input image from the set of few-shot candidates.
We do not filter training data based on the question type.
In the oracle-style setting of the Shared Task, it would have been possible to additionally filter based on question type. We do so only indirectly by searching for questions and images with high embedding similarity, which makes our approach more directly applicable to real-world scenarios, where the question type may not be provided. Moreover, the question type \enquote{unanswerable} directly reveals the gold answer.


Our retrieval method ranks instances. It can thus be used to retrieve an arbitrary number of few-shot examples.
We evaluate the performance of these retrieval strategies in one-shot and two-shot settings.
When using two examples, the model is given one answerable and one unanswerable example.



\section{Development Results and Ablations}
Since our approach does not require any fine-tuning, we combine the training and validation sets into one development set. This section describes our careful experimentation and ablation studies on the development set.
For more information on the dataset, see Appendix \ref{ssec:dataset_characteristics}.

We rely on the metrics of the Shared Task, F1, Precision, and Recall of ROUGE-1, ROUGE-L \citep{lin-2004-rouge}, and BERTScore \citep[BERTS,][]{Zhang2020BERTScore}, respectively, to evaluate our approach.
However, we focus on ROUGE-1 F1, as the BERTS scores are similar for all approaches, and the ROUGE-L scores are comparable to ROUGE-1.

We run our experiments on Nvidia A100 (80 GB) GPUs, using up to 4 GPUs in parallel. The total amount of GPU hours was about 3600h.

\begin{figure}[t]
\includegraphics[width=1\columnwidth]{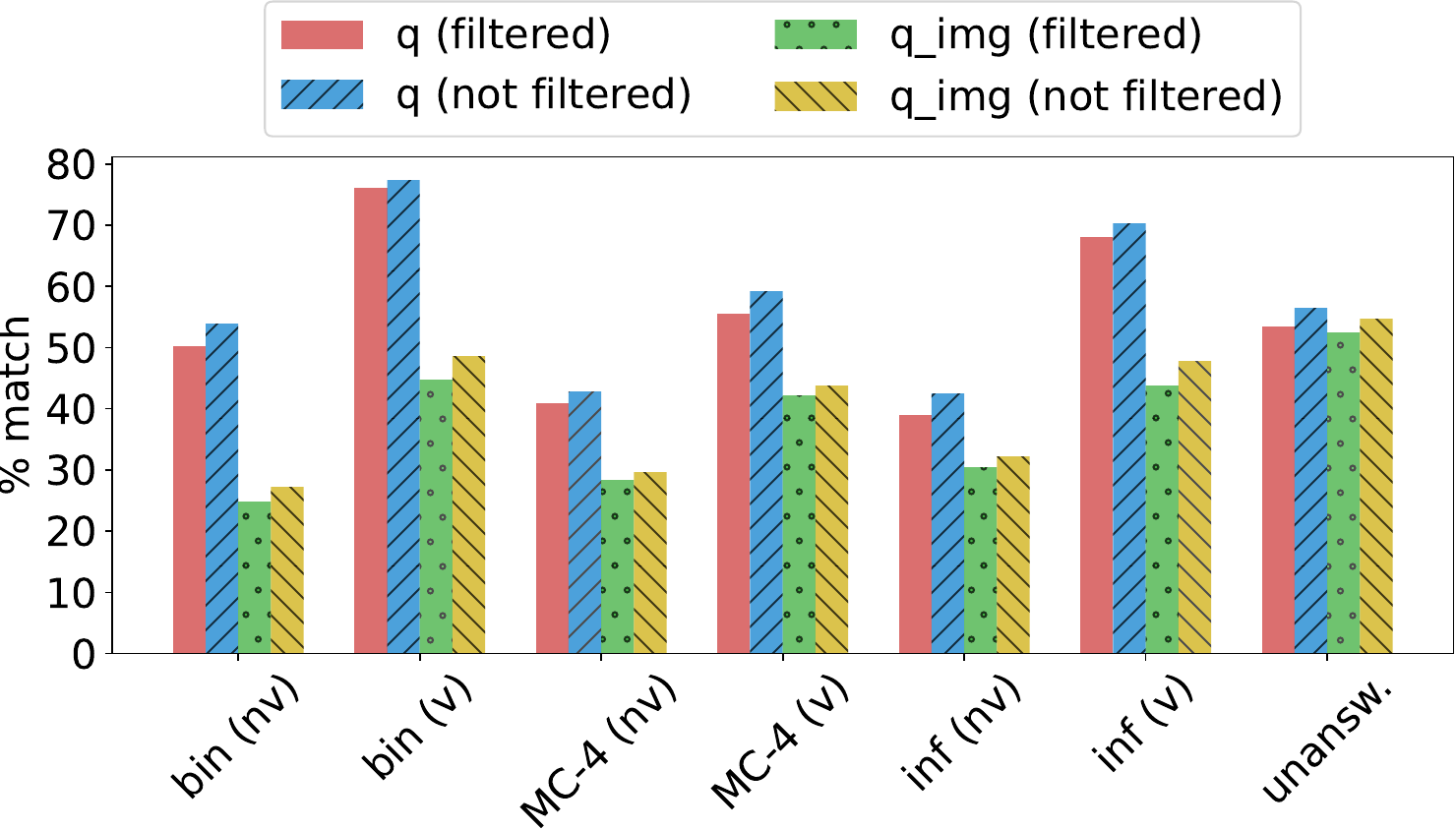}
  \caption {Percentage of selected one-shot example matching the question type of the input instance. bin = binary question, MC-4 = four answer options, inf = infinite answer set, unansw. = unanswerable, (v) = visual, (nv) = non-visual, filtered = filter for same figure type, not filtered = search in entire train set, q = question similarity, img = image similarity.} 
  \label{fig:acc_question_type}
\end{figure}

\subsection{Retrieval of Few-Shot Examples}
\label{sec:ret_few_shot_ex}
As shown in \autoref{fig:acc_question_type}, the degree to which the question type of the retrieved few-shot examples matches that of the input instance varies greatly by question type.
Searching for examples using only question similarity leads to matching the input instance's question type far more often than searching using image and question similarity.
However, this does not seem to make a marked difference in overall performance.
We found BLIP-2's text-image embeddings to primarily reflect the image content, resulting in many ties.\footnote{Our tie-breaking strategy leads to instances of the question type \enquote{closed-ended infinite answer set visual} to be selected, which comes first in the training set ordering for each image.}

\subsection{Impact of Few-Shot Examples}

\begin{table*}
\footnotesize
\centering
\setlength\tabcolsep{4pt}
\begin{tabular}{l|l|rrr|rrr|rrr}
\toprule
\textbf{Setting} & \textbf{Configuration} & R1-F1 & R1-P & R1-R & RL-F1 & RL-P & RL-R & BS-F1 & BS-P & BS-R \\
\midrule
Individual runs & InternVL (0s) & 74.2 & 75.2 & 74.9 & 74.1 & 75.1 & 74.8 & 97.1 & 97.3 & 97.0 \\
(dev set) & InternVL (1s\_q\_f) & 74.7 & 75.7 & 74.8 & 74.6 & 75.6 & 74.8 & 97.8 & 97.9 & 97.8 \\
 & InternVL (1s\_q\_nf) & 74.5 & 75.5 & 74.6 & 74.4 & 75.4 & 74.6 & 97.8 & 97.8 & 97.7 \\
 & InternVL (1s\_q\_img\_f) & 74.8 & 75.6 & \textbf{75.2} & 74.7 & 75.6 & \textbf{75.1} & 97.8 & 97.8 & 97.8 \\
 & InternVL (1s\_q\_img\_nf) & 74.7 & 75.7 & 75.1 & 74.6 & 75.6 & 75.0 & 97.8 & 97.8 & 97.8 \\
 & InternVL (1s\_q\_img\_f, BLIP2) & \textbf{75.0} & \textbf{76.0} & \textbf{75.2} & \textbf{74.9} & \textbf{76.0} & \textbf{75.1} & \textbf{97.9} & \textbf{98.0} & \textbf{97.9} \\
 & Pixtral (0s) & 71.4 & 72.5 & 72.4 & 71.2 & 72.4 & 72.2 & 96.3 & 96.6 & 96.0 \\
 & Pixtral (1s\_q\_f) & 72.8 & 74.0 & 73.1 & 72.7 & 73.9 & 73.0 & 97.5 & 97.6 & 97.5 \\
 & Pixtral (1s\_q\_nf) & 72.3 & 73.5 & 72.6 & 72.2 & 73.4 & 72.5 & 97.5 & 97.5 & 97.5 \\
 & Pixtral (1s\_q\_img\_f) & 72.8 & 74.1 & 73.1 & 72.7 & 74.0 & 73.0 & 97.4 & 97.5 & 97.3 \\
 & Pixtral (1s\_q\_img\_nf) & 72.8 & 74.0 & 73.2 & 72.7 & 73.9 & 73.1 & 97.4 & 97.5 & 97.3 \\
 & Pixtral (2s\_q\_f) & 73.9 & 75.2 & 74.0 & 73.8 & 75.1 & 73.9 & 97.7 & 97.8 & 97.7 \\
 & Pixtral (2s\_q\_nf) & 73.7 & 75.0 & 73.8 & 73.6 & 74.9 & 73.7 & 97.7 & 97.8 & 97.7 \\
 & Pixtral (2s\_q\_img\_f) & 74.1 & 75.5 & 74.2 & 74.0 & 75.4 & 74.1 & 97.7 & 97.8 & 97.6 \\
 & Pixtral (2s\_q\_img\_nf) & 73.8 & 75.2 & 73.9 & 73.7 & 75.1 & 73.8 & 97.6 & 97.8 & 97.6 \\
 \midrule
Ensembles& Question/Figure-Type Ensemble & 76.6 & 78.0 & 76.5 & 76.5 & 77.8 & 76.4 & 97.9 & 98.0 & \textbf{97.9} \\
 (dev set) & Confidence-Informed Ensemble & \textbf{76.9} & \textbf{78.2} & \textbf{76.8} & \textbf{76.8} & \textbf{78.1} & \textbf{76.8} & \textbf{98.0} & \textbf{98.1} & \textbf{97.9} \\
 \midrule
Results on & InvernVL (1s\_q\_img\_f, BLIP2) & 77.2 & 78.0 & 77.4 & 77.2 & 77.9 & 77.3 & 98.1 & 98.2 & \textbf{98.1}\\
test set &  Question/Figure-Type Ensemble & 77.7 & 78.8 & 77.7 & 77.6 & 78.7 & 77.6 & 98.1 & 98.2 & 98.0 \\
& Confidence-Informed Ensemble & \textbf{78.6} & \textbf{79.7} & \textbf{78.6} & \textbf{78.6} & \textbf{79.6} & \textbf{78.5} & \textbf{98.2} & \textbf{98.3} & \textbf{98.1} \\
\bottomrule
\end{tabular}
  \caption{Results of individual runs vs.~ensembles on development and test set. Abbreviations for few-shot example selection: \#s = \#-shot, q = question similarity, img = image similarity, f = filter for same figure type, nf = no filtering (search in entire train set). Metrics: R1 = ROUGE-1, RL = ROUGE-L, BS = BERTS, P = Precision, R = Recall. Question/Figure-Type Ensemble refers to the approach described in section \ref{sec:q_fig_type_ens} and Confidence-Informed Ensemble to that of section \ref{sec:conf_ens}.}
  \label{tab:main_res}
\end{table*}

\begin{figure}[t]
 \includegraphics[width=1\columnwidth]{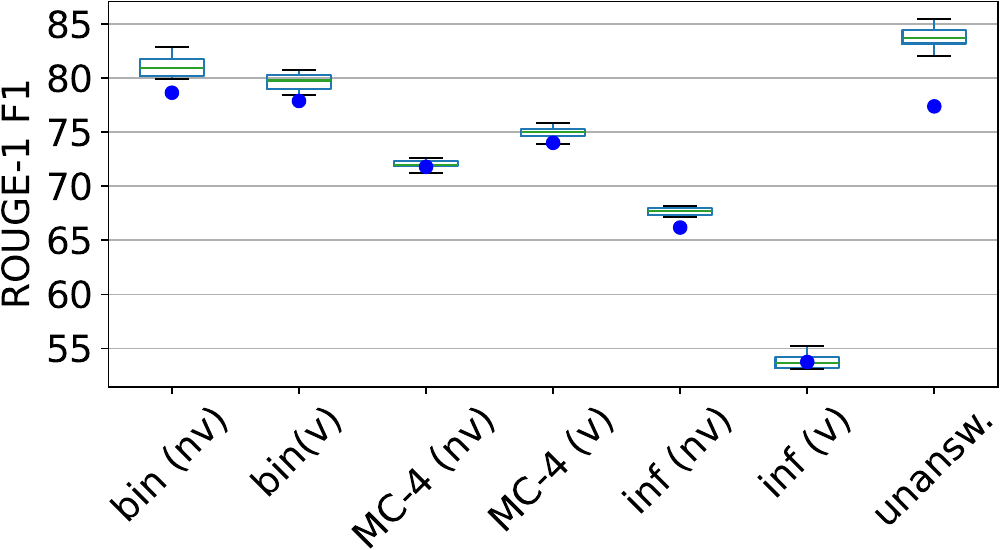}
  \caption {ROUGE-1 F1 scores per question type. Boxplot: 1-shot and 2-shot question and question+image similarity configurations of Pixtral-Large-2411.
  Blue dots = Pixtral-Large-2411 (0-shot).
  } 
  \label{fig:f1_qa_type_boxplot}
\end{figure}

\begin{figure}[t]
 \includegraphics[width=1\columnwidth]{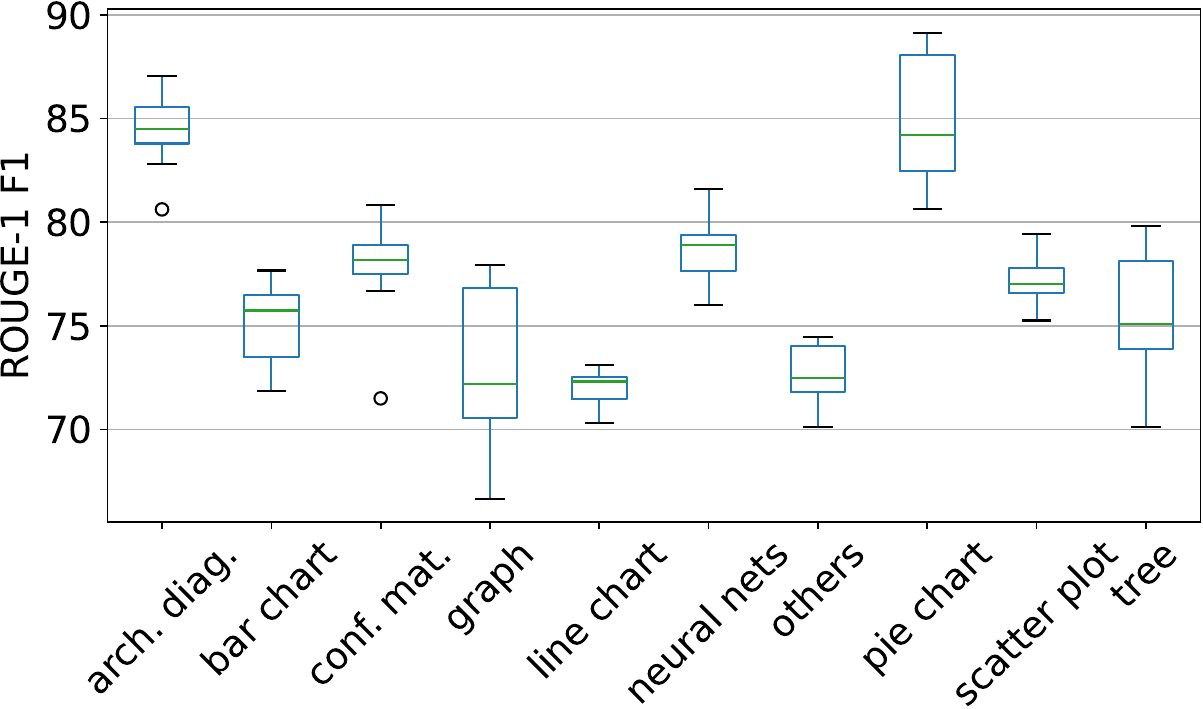}
  \caption {ROUGE-1 F1 scores per figure type of all configurations (zero- and few-shot) of both MLLMs visualized as boxplots.} 
  \label{fig:f1_fig_type_boxplot}
\end{figure}

\begin{table}[t]
\footnotesize
\centering
\setlength\tabcolsep{2pt}
\begin{tabular}{l|r}
\toprule
\textbf{Approach} & \textbf{Precision} \\
\midrule
Pixtral (0s) & 93.0\\
Pixtral (1s\_q\_f) & 89.2\\
Pixtral (1s\_q\_img\_f) & 89.3\\
Pixtral (1s\_q\_img\_nf) & 90.3\\
Pixtral (1s\_q\_nf) & 88.7\\
Pixtral (2s\_q\_f) & 92.7\\
Pixtral (2s\_q\_img\_f) & 94.1\\
Pixtral (2s\_q\_img\_nf) & 93.7\\
Pixtral (2s\_q\_nf) & 93.3\\
\bottomrule
\end{tabular}
  \caption{Precision of instances predicted to be unanswerable.}
  \label{tab:prec_unanswerable}
\end{table}

\autoref{tab:main_res} compares the effectiveness of 
InternVL3-78B and Pixtral-Large-2411
using various few-shot settings with that of our ensemble approaches.
Adding few-shot examples generally improves performance. 
We cannot report 2-shot results for InternVL3-78B because its context window is too small to incorporate two examples. 
Comparing performance by question type reveals that adding examples can be highly beneficial, e.g., for recognizing unanswerable questions, though they can also be distracting (see \autoref{fig:f1_qa_type_boxplot} or Appendix \ref{ssec:detailed_res_appendix}).
However, adding two examples is almost always beneficial. 
Furthermore, using one answerable and one unanswerable example helps the model to distinguish between these two types of instances, especially when compared to using only one example (see \autoref{tab:prec_unanswerable}). 

\subsection{Question/Figure Type Ensemble}
\label{sec:q_fig_type_ens}
To determine the best configuration of MLLM and few-shot strategy for each pair of question type and figure type, we systematically search for the optimal ensemble settings by obtaining and analyzing distributions of performance scores over subsets of the data similar to cross-validation.

While there appears to be a general trend of enhanced performance with the use of examples (see \autoref{tab:main_res}), our findings reveal considerable variations in the performance of our configurations across different question and figure types (see \autoref{fig:f1_qa_type_boxplot} and \autoref{fig:f1_fig_type_boxplot} or Appendix \ref{ssec:detailed_res_appendix}).
Therefore, we use the results on the development set to systematically identify the optimal combination of configurations that work well across as many subsets of the data as possible. 
The dataset consists of seven evenly represented question types and various figure types that are not evenly distributed.
We record performance scores for each figure type separately.
To avoid overfitting, we summarize all figure types that encompass less than two percent of the total number of figures into the figure type \enquote{others}, which leads to nine groups with homogeneous figure types (line chart, tree, scatter plot, pie chart, bar chart, architecture diagram, neural networks, confusion matrix, graph) plus one group of the \enquote{others}, i.e., 10 groups in total.
For the largest figure type, i.e., line chart, we divide the data into seven groups by further dividing the data by question type.
In total, we divide the data into 16 groups (8 homogeneous figure types, 1 \enquote{others}, and 7 subsets with line charts).


We split the data of each group into 5 folds and compute performance scores.
We repeat this process
at least 10 times with different splits until the predicted best-performing configuration remains constant.
In each fold, we calculate the ROUGE-1 F1 score for all configurations, then subtract the highest score achieved in that fold. For each configuration, we then compute the mean of these scores across all folds and all runs. 
The best-performing configuration is identified by the highest score.
For the final chosen configuration of this ensemble, refer to \autoref{tab:best_comb_cross_val} in Appendix \ref{ssec:detailed_res_appendix}.

\subsection{Confidence-Informed Ensemble}
\label{sec:conf_ens}
\begin{figure}[t]
 \includegraphics[width=1\columnwidth]{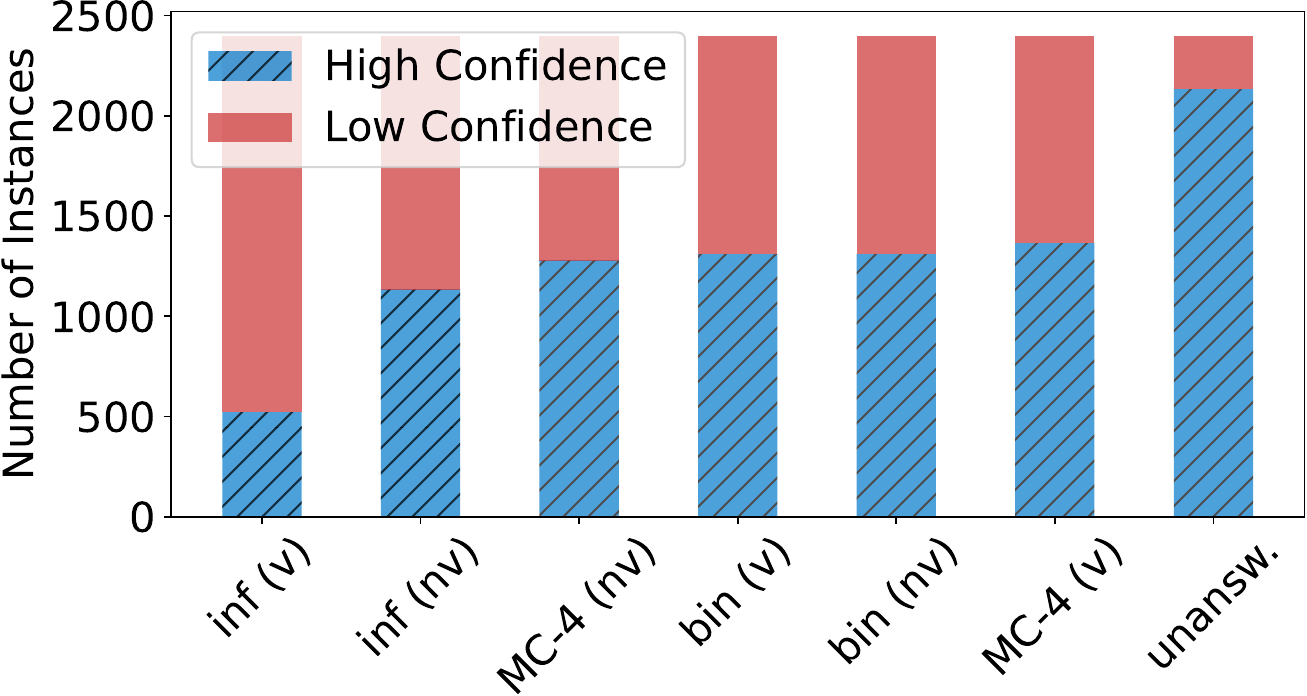}
  \caption {Number of instances having received high confidence answer of InternVL3-78B (1s\_q\_img\_f, BLIP) by question type.} 
  \label{fig:num_high_conf_pred}
\end{figure}

\begin{figure}[t]
 \centering
 \includegraphics[width=0.7\columnwidth]{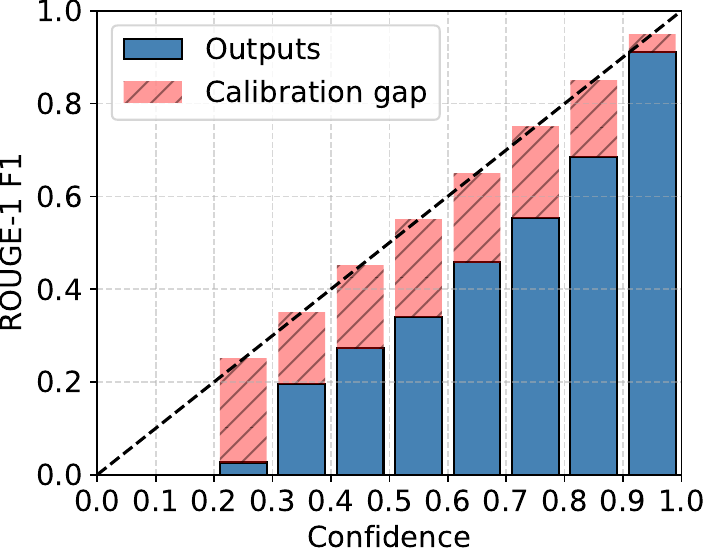}
  \caption {Calibration plot for InternVL3-78B (1s\_q\_img\_f, BLIP) showing that instances with confidence score $\geq0.9$ have high expected accuracy.} 
  \label{fig:conf_intern_blip}
\end{figure}


\autoref{fig:conf_intern_blip} shows that InternVL3-78B (1s\_q\_img\_f, BLIP2) with examples derived from BLIP-2, which focus primarily on image similarity as explained in Sec.~\ref{sec:ret_few_shot_ex}, is meaningfully calibrated. 
This means that high confidence scores indicate highly likely correct instances (refer to Appendix \ref{ssec:detailed_res_appendix} for detailed results). 
Thus, for our final submission, we directly use all predictions from this model with a confidence score of at least 90\%, which corresponds to approximately half of the instances, in the initial stage. 
As shown in \autoref{fig:num_high_conf_pred}, the number of high-confidence instances varies by question type.
The model is most confident on identifying unanswerable questions, while it is least sure about its answers for questions with infinite answers sets about the image's visual features.
After removing high-confidence instances, the performance per question type varies widely between our configruations (see Appendix \ref{ssec:detailed_res_appendix} for detailed results). 
The best configuration per question type does not seem to depend on whether the question incorporates visual or non-visual features.
Since the vast majority of the remaining instances are of figure type \textit{line chart}, we do not perform cross-validation to determine the optimal configuration of approaches. 
Instead, we select the best-performing approach for each question type, while also trying to reduce the number of approaches required.

As can be seen in \autoref{fig:system_overview}, we use the following models for the remaining instances: Pixtral-Large-2411 (2s\_q\_f) for binary questions, Pixtral-Large-2411 (2s\_q\_img\_f) for questions with an infinite answer set, and InternVL3-78B (1s\_q\_f) for all others.

\section{Results on Test Set} 
\autoref{tab:main_res} also shows the results of our approaches on the test set, indicating that our ensembling strategies improve the performance compared to using only one approach to answer all questions.
On the test set, we also find the confidence-informed ensemble to work best, while the question/figure type ensemble outperforms the simple InternVL model not as strongly as on the development set.
The confidence-informed ensemble is the approach submitted for the leaderboard, ranking third in the official evaluation (almost on par with the second-ranking system) as shown in \autoref{tab:leaderboard_scores}, and outperforming the baseline by about 4 percentage points.

\section{Discussion and Conclusion}
This paper described our submission to the SciVQA 2025 Shared Task. 
Our results show that MLLMs are highly effective at answering questions about scientific figures. However, performance varies greatly by question type. Results on finite answer sets are considerably better than on infinite ones. 
In particular, answering infinite answer set questions about visual features of images remains challenging, highlighting the need for a more sophisticated approach. 

The use of few-shot examples 
improves performance. However, there are no major performance differences between retrieving the examples by question or question-image similarity. 

\section*{Limitations}
Since the ACL-Fig and SciGraphQA datasets, on which the figures in this Shared Task are based, rely on images published several years ago, some of 
these images may have already been exposed to MLLMs during training.

Another limitation is performance on unanswerable questions. Although our approach performed best on this question type, it is difficult to determine if it would perform equally well on real-world unanswerable questions. This is because the unanswerable questions in this dataset follow a different pattern than the answerable ones. For example, they mostly refer to material unavailable to the model and often do not focus on the images' visual/non-visual features.



\section*{Acknowledgments}
The authors gratefully acknowledge the scientific support and HPC resources provided by the Erlangen National High Performance Computing Center (NHR@FAU) of the Friedrich-Alexander-Universität Erlangen-Nürnberg (FAU) under the BayernKI project v110ee. BayernKI funding is provided by Bavarian state authorities.


\input{acl_latex.bbl}
\appendix

\section{Appendix}

\subsection{Detailed Results on Development Set}
\label{ssec:detailed_res_appendix}

\autoref{tab:res_by_question_type} shows the detailed results of the different zero- and few-shot approaches on the development set, broken down by question type. Performance varies greatly between question types, indicating that questions with an infinite answer set are more difficult. Furthermore, performance depends on the MLLM and few-shot configuration used. Mostly, using examples is beneficial for performance.

As shown in \autoref{tab:res_by_figure_type}, the performance of the different configurations also depends on the figure type of the image.

\autoref{tab:conf_scores_counts} reports the ROUGE-1 F1 score per confidence bin and the relative proportion of respective bin of the development set. 
Interestingly, the configuration that uses BLIP-2 to retrieve similar examples is well-calibrated for high confidence. In general, InternVL3-78B appears to be better calibrated than Pixtral-Large-2411 for our task.

The performance of different approaches per question type can be seen in \autoref{tab:res_by_qa_type_without_high_conf} after having removed all instances of InternvL3-78B (1s\_q\_img\_f, BLIP) with a confidence of $\geq 90 \%$. 
Performance is worse compared to \autoref{tab:res_by_question_type} since the high confidence answers are removed.
Nevertheless, there are still large performance differences between the different approaches.

\autoref{tab:best_comb_cross_val} shows the best configurations per figure and question type identified via cross-validation for the Question/Figure Type Ensemble.

\begin{table*}
\footnotesize
\centering
\setlength\tabcolsep{2pt}
\begin{tabular}{l|rrrrrrr}
\toprule
Approach & binary (nv) & binary(v) & MC-4 (nv) & MC-4 (v) & inf (nv) & inf (v) & unanswerable \\
\midrule
InternVL (0s) & 78.9 & 80.9 & 76.6 & 79.3 & 66.4 & 51.2 & 86.2 \\
InternVL (1s\_q\_f) & 82.0 & 80.8 & \textbf{76.9} & \textbf{79.9} & 63.4 & 49.3 & 90.4 \\
InternVL (1s\_q\_nf) & 82.0 & 81.0 & 76.3 & 79.1 & 63.5 & 50.0 & 89.5 \\
InternVL (1s\_q\_img\_f) & 80.4 & \textbf{81.3} & 76.3 & 79.3 & 65.3 & 49.5 & \textbf{91.2} \\
InternVL (1s\_q\_img\_nf) & 80.6 & 80.8 & 76.3 & 78.6 & 65.7 & 50.1 & 91.0 \\
InternVL (1s\_q\_img\_f, BLIP) & 82.7 & 80.6 & 76.7 & 79.0 & 67.4 & 51.8 & 87.0 \\
Pixtral (0s) & 78.6 & 77.9 & 71.8 & 74.0 & 66.2 & 53.8 & 77.4 \\
Pixtral (1s\_q\_f) & 80.2 & 79.5 & 71.7 & 73.9 & 67.4 & 53.4 & 83.4 \\
Pixtral (1s\_q\_nf) & 79.9 & 78.5 & 71.2 & 74.1 & 67.5 & 53.1 & 82.0 \\
Pixtral (1s\_q\_img\_f) & 80.3 & 79.0 & 71.9 & 75.0 & 67.1 & 53.1 & 83.2 \\
Pixtral (1s\_q\_img\_nf) & 79.9 & 79.0 & 72.3 & 75.1 & 67.3 & 53.2 & 83.0 \\
Pixtral (2s\_q\_f) & \textbf{82.9} & 80.7 & 72.1 & 74.9 & \textbf{68.1} & 54.0 & 84.4 \\
Pixtral (2s\_q\_nf) & 82.1 & 80.3 & 71.9 & 75.0 & 68.0 & 54.6 & 84.0 \\
Pixtral (2s\_q\_img\_f) & 81.5 & 80.4 & 72.6 & 75.8 & 67.9 & \textbf{55.2} & 85.4 \\
Pixtral (2s\_q\_img\_nf) & 81.6 & 80.0 & 72.4 & 75.9 & \textbf{68.1} & 54.1 & 84.5 \\
\bottomrule
\end{tabular}
  \caption{Results (ROUGE-1 F1 scores) on development set by question type. v=visual, nv= non-visual.}
  \label{tab:res_by_question_type}
\end{table*}


\begin{table*}
\footnotesize
\centering
\setlength\tabcolsep{2pt}
\begin{tabular}{l|rrrrrrrrrr}
\toprule
\makecell{Approach} & \makecell{architecture\\diagram} & \makecell{bar\\chart} & \makecell{confusion\\matrix} & \makecell{graph} & \makecell{line\\chart} & \makecell{neural\\networks} & \makecell{others} & \makecell{pie\\chart} & \makecell{scatter\\plot} & \makecell{tree} \\
\midrule
InternVL (0s) & 83.9 & \textbf{77.7} & 77.0 & 75.4 & 72.2 & 78.9 & 73.4 & 87.4 & 78.0 & 76.2 \\
InternVL (1s\_q\_f) & 86.2 & 77.3 & 76.7 & 76.8 & 72.5 & 79.2 & \textbf{74.5} & \textbf{89.2} & 75.6 & 78.6 \\
InternVL (1s\_q\_nf) & 85.0 & 76.4 & 78.6 & 77.4 & 72.5 & 79.2 & 73.4 & 88.0 & 76.2 & 77.8 \\
InternVL (1s\_q\_img\_f) & 86.2 & 76.6 & 78.4 & 77.7 & 72.6 & 78.8 & 74.4 & 88.7 & 77.1 & 78.2 \\
InternVL (1s\_q\_img\_nf) & 85.4 & 76.1 & 78.3 & \textbf{78.0} & 72.6 & 79.7 & 73.9 & 88.0 & 76.9 & \textbf{79.2} \\
InternVL (1s\_q\_img\_f, BLIP) & \textbf{87.0} & 76.3 & 78.8 & 77.1 & \textbf{72.8} & \textbf{81.6} & 74.4 & 88.1 & 77.9 & 78.1 \\
Pixtral (0s) & 80.6 & 72.8 & 71.5 & 66.7 & 70.3 & 76.0 & 70.1 & 82.8 & 75.3 & 70.1 \\
Pixtral (1s\_q\_f) & 84.7 & 73.8 & 77.5 & 67.9 & 71.3 & 77.5 & 72.0 & 82.1 & 76.7 & 73.5 \\
Pixtral (1s\_q\_nf) & 83.3 & 73.4 & 77.7 & 69.8 & 71.0 & 77.1 & 70.3 & 81.6 & 75.7 & 72.6 \\
Pixtral (1s\_q\_img\_f) & 82.8 & 73.3 & 78.1 & 70.7 & 71.4 & 77.7 & 71.4 & 80.9 & 76.9 & 73.9 \\
Pixtral (1s\_q\_img\_nf) & 84.1 & 71.8 & 78.0 & 70.3 & 71.5 & 76.2 & 71.6 & 80.6 & 76.9 & 74.3 \\
Pixtral (2s\_q\_f) & 83.6 & 76.2 & 79.9 & 71.4 & 72.3 & 79.6 & 72.7 & 82.6 & 77.4 & 74.5 \\
Pixtral (2s\_q\_nf) & 83.9 & 73.5 & 77.4 & 72.6 & 72.3 & 78.6 & 72.3 & 83.7 & 77.8 & 74.2 \\
Pixtral (2s\_q\_img\_f) & 85.2 & 75.4 & 80.4 & 71.7 & 72.5 & 78.9 & 71.9 & 84.4 & \textbf{78.9} & 75.7 \\
Pixtral (2s\_q\_img\_nf) & 84.3 & 74.2 & \textbf{80.8} & 71.3 & 72.3 & 79.3 & 72.2 & 84.0 & 77.6 & 73.8 \\
\bottomrule
\end{tabular}
  \caption{Results (ROUGE-1 F1 scores) on development set by figure type.}
  \label{tab:res_by_figure_type}
\end{table*}

\begin{table*}
\footnotesize
\centering
\setlength\tabcolsep{2pt}
\begin{tabular}{l|rrrrrrr}
\toprule
Approach & 0.3\_0.4 & 0.4\_0.5 & 0.5\_0.6 & 0.6\_0.7 & 0.7\_0.8 & 0.8\_0.9 & 0.9\_1.0 \\
\midrule
InternVL (0s) & 15.2 (0.1) & \textbf{34.5} (1.2) & \textbf{48.7} (5.3) & \textbf{55.4} (13.6) & \textbf{69.9} (21.2) & \textbf{73.3} (18.2) & 87.9 (40.4) \\
InternVL (1s\_q\_f) & 18.5 (0.1) & 30.2 (0.8) & 30.3 (2.9) & 45.4 (7.4) & 56.8 (14.1) & 66.4 (16.2) & 88.0 (58.5) \\
InternVL (1s\_q\_nf) & 22.9 (0.1) & 25.0 (0.9) & 31.1 (3.1) & 45.5 (7.2) & 55.5 (14.1) & 66.3 (16.2) & 88.1 (58.4) \\
InternVL (1s\_q\_img\_f) & 21.2 (0.1) & 30.9 (0.8) & 37.6 (3.4) & 48.8 (8.4) & 57.6 (14.8) & 68.6 (16.5) & 88.1 (55.9) \\
InternVL (1s\_q\_img\_nf) & 15.9 (0.1) & 28.3 (0.8) & 38.7 (3.3) & 48.1 (8.4) & 56.1 (14.7) & 69.2 (16.7) & 88.3 (55.9) \\
InternVL (1s\_q\_img\_f, BLIP) & 19.6 (0.2) & 27.3 (1.0) & 34.0 (3.5) & 45.8 (8.3) & 55.4 (15.3) & 68.6 (17.8) & \textbf{91.1} (53.9) \\
Pixtral (0s) & \textbf{25.0} (0.0) & 24.7 (0.3) & 34.6 (1.8) & 45.4 (6.0) & 61.2 (17.3) & 64.5 (24.6) & 83.0 (50.0) \\
Pixtral (1s\_q\_f) & 0.0 (0.0) & 31.1 (0.3) & 33.7 (1.6) & 43.3 (5.6) & 54.6 (13.4) & 61.8 (19.9) & 84.7 (59.2) \\
Pixtral (1s\_q\_nf) & 11.4 (0.0) & 32.6 (0.3) & 32.8 (1.6) & 42.1 (5.6) & 52.9 (13.5) & 61.7 (19.9) & 84.5 (59.1) \\
Pixtral (1s\_q\_img\_f) & 16.3 (0.0) & 25.8 (0.3) & 39.5 (1.8) & 45.5 (6.2) & 55.9 (13.8) & 62.9 (20.8) & 84.8 (57.1) \\
Pixtral (1s\_q\_img\_nf) & 19.0 (0.0) & 23.5 (0.3) & 40.0 (1.7) & 46.9 (6.2) & 54.3 (14.0) & 63.2 (20.5) & 85.0 (57.1) \\
Pixtral (2s\_q\_f) & 0.0 (0.0) & 24.1 (0.2) & 37.8 (1.2) & 40.8 (4.6) & 52.8 (12.3) & 62.1 (19.2) & 84.9 (62.5) \\
Pixtral (2s\_q\_nf) & 0.0 (0.0) & 20.1 (0.2) & 31.4 (1.2) & 42.3 (4.6) & 52.3 (11.8) & 60.9 (19.6) & 85.0 (62.6) \\
Pixtral (2s\_q\_img\_f) & 14.3 (0.0) & 25.6 (0.1) & 33.6 (1.4) & 47.2 (5.2) & 55.1 (12.6) & 61.9 (20.3) & 85.6 (60.4) \\
Pixtral (2s\_q\_img\_nf) & 0.0 (0.0) & 28.8 (0.2) & 34.2 (1.4) & 42.8 (5.1) & 52.6 (12.8) & 63.1 (19.9) & 85.4 (60.7) \\
\bottomrule
\end{tabular}
  \caption{Results (ROUGE-1 F1 scores) per confidence bin. The values in brackets indicate the relative proportion of instances in each bin.}
  \label{tab:conf_scores_counts}
\end{table*}

\begin{table*}
\footnotesize
\centering
\setlength\tabcolsep{2pt}
\begin{tabular}{l|rrrrrrr}
\toprule
Approach & binary (nv) & binary(v) & MC-4 (nv) & MC-4 (v) & inf (nv) & inf (v) & unanswerable \\
\midrule
InternVL (0s) & 64.8 & 68.8 & 60.6 & 62.0 & 50.6 & 45.0 & 34.3 \\
InternVL (1s\_q\_f) & 70.7 & 68.3 & \textbf{61.0} & \textbf{63.1} & 46.9 & 42.0 & 48.6 \\
InternVL (1s\_q\_img\_f) & 67.8 & 69.2 & 60.1 & 62.2 & 49.9 & 42.3 & 50.2 \\
InternVL (1s\_q\_img\_nf) & 68.2 & 67.9 & 59.6 & 60.8 & 50.1 & 43.0 & \textbf{50.8} \\
InternVL (1s\_q\_nf) & 70.2 & 68.5 & 60.4 & 62.0 & 47.3 & 42.8 & 42.8 \\
Pixtral (0s) & 67.8 & 67.1 & 58.1 & 58.4 & 52.3 & 48.1 & 29.8 \\
Pixtral (1s\_q\_f) & 70.3 & 68.9 & 57.2 & 57.7 & 53.8 & 48.0 & 38.2 \\
Pixtral (1s\_q\_img\_f) & 71.4 & 69.0 & 57.6 & 59.2 & 53.8 & 48.0 & 32.2 \\
Pixtral (1s\_q\_img\_nf) & 70.3 & 68.0 & 58.8 & 59.3 & 53.8 & 48.1 & 33.1 \\
Pixtral (1s\_q\_nf) & 70.1 & 67.0 & 56.8 & 58.1 & 53.6 & 47.8 & 34.7 \\
Pixtral (2s\_q\_f) & \textbf{74.0} & \textbf{70.8} & 57.5 & 58.9 & 54.8 & 48.4 & 36.8 \\
Pixtral (2s\_q\_img\_f) & 72.5 & 69.4 & 58.2 & 59.9 & 55.2 & \textbf{49.5} & 39.0 \\
Pixtral (2s\_q\_img\_nf) & 72.7 & 69.0 & 58.4 & 60.0 & \textbf{55.4} & 48.4 & 35.8 \\
Pixtral (2s\_q\_nf) & 73.2 & 69.9 & 57.4 & 59.0 & 54.5 & \textbf{49.5} & 37.0 \\
\bottomrule
\end{tabular}
  \caption{Results (ROUGE-1 F1 scores) on development set by question type after removing high confidence instances of run with InternVL3-78B (1s\_q\_img\_f with BLIP-2).}
  \label{tab:res_by_qa_type_without_high_conf}
\end{table*}

\begin{table*}
\footnotesize
\centering
\setlength\tabcolsep{2pt}
\begin{tabular}{l|ccccccc}
\toprule
Figure Type & inf (v) & inf (nv) & bin (v) & bin (nv) & MC-4 (v) & MC-4 (nv) & unansw. \\
\midrule
\makecell{line chart} & \makecell{Pixtral \\(2s\_q\_img\_f)} & \makecell{Pixtral \\(2s\_q\_nf)} & \makecell{Pixtral \\(2s\_q\_f)} & \makecell{Pixtral \\(2s\_q\_f)} & \makecell{InternVL \\(1s\_q\_f)} & \makecell{InternVL \\(1s\_q\_f)} & \makecell{InternVL \\(1s\_q\_img\_f)} \\
\makecell{tree} & \makecell{InternVL \\(1s\_q\_img\_nf)} & \makecell{InternVL \\(1s\_q\_img\_nf)} & \makecell{InternVL \\(1s\_q\_img\_nf)} & \makecell{InternVL \\(1s\_q\_img\_nf)} & \makecell{InternVL \\(1s\_q\_img\_nf)} & \makecell{InternVL \\(1s\_q\_img\_nf)} & \makecell{InternVL \\(1s\_q\_img\_nf)} \\
\makecell{scatter plot} & \makecell{Pixtral \\(2s\_q\_img\_f)} & \makecell{Pixtral \\(2s\_q\_img\_f)} & \makecell{Pixtral \\(2s\_q\_img\_f)} & \makecell{Pixtral \\(2s\_q\_img\_f)} & \makecell{Pixtral \\(2s\_q\_img\_f)} & \makecell{Pixtral \\(2s\_q\_img\_f)} & \makecell{Pixtral \\(2s\_q\_img\_f)} \\
\makecell{pie chart} & \makecell{InternVL \\(1s\_q\_f)} & \makecell{InternVL \\(1s\_q\_f)} & \makecell{InternVL \\(1s\_q\_f)} & \makecell{InternVL \\(1s\_q\_f)} & \makecell{InternVL \\(1s\_q\_f)} & \makecell{InternVL \\(1s\_q\_f)} & \makecell{InternVL \\(1s\_q\_f)} \\
\makecell{bar chart} & \makecell{InternVL \\(0s)} & \makecell{InternVL \\(0s)} & \makecell{InternVL \\(0s)} & \makecell{InternVL \\(0s)} & \makecell{InternVL \\(0s)} & \makecell{InternVL \\(0s)} & \makecell{InternVL \\(0s)} \\
\makecell{architecture\\diagram} & \makecell{InternVL \\(1s\_q\_f)} & \makecell{InternVL \\(1s\_q\_f)} & \makecell{InternVL \\(1s\_q\_f)} & \makecell{InternVL \\(1s\_q\_f)} & \makecell{InternVL \\(1s\_q\_f)} & \makecell{InternVL \\(1s\_q\_f)} & \makecell{InternVL \\(1s\_q\_f)} \\
\makecell{neural\\networks} & \makecell{InternVL \\(1s\_q\_img\_nf)} & \makecell{InternVL \\(1s\_q\_img\_nf)} & \makecell{InternVL \\(1s\_q\_img\_nf)} & \makecell{InternVL \\(1s\_q\_img\_nf)} & \makecell{InternVL \\(1s\_q\_img\_nf)} & \makecell{InternVL \\(1s\_q\_img\_nf)} & \makecell{InternVL \\(1s\_q\_img\_nf)} \\
\makecell{confusion\\matrix} & \makecell{Pixtral \\(2s\_q\_img\_nf)} & \makecell{Pixtral \\(2s\_q\_img\_nf)} & \makecell{Pixtral \\(2s\_q\_img\_nf)} & \makecell{Pixtral \\(2s\_q\_img\_nf)} & \makecell{Pixtral \\(2s\_q\_img\_nf)} & \makecell{Pixtral \\(2s\_q\_img\_nf)} & \makecell{Pixtral \\(2s\_q\_img\_nf)} \\
\makecell{graph} & \makecell{InternVL \\(1s\_q\_img\_nf)} & \makecell{InternVL \\(1s\_q\_img\_nf)} & \makecell{InternVL \\(1s\_q\_img\_nf)} & \makecell{InternVL \\(1s\_q\_img\_nf)} & \makecell{InternVL \\(1s\_q\_img\_nf)} & \makecell{InternVL \\(1s\_q\_img\_nf)} & \makecell{InternVL \\(1s\_q\_img\_nf)} \\
\makecell{others} & \makecell{InternVL \\(1s\_q\_f)} & \makecell{InternVL \\(1s\_q\_f)} & \makecell{InternVL \\(1s\_q\_f)} & \makecell{InternVL \\(1s\_q\_f)} & \makecell{InternVL \\(1s\_q\_f)} & \makecell{InternVL \\(1s\_q\_f)} & \makecell{InternVL \\(1s\_q\_f)} \\
\bottomrule
\end{tabular}
  \caption{Best configurations for combination of figure type and question type identified via cross-validation for Question/Figure Type Ensemble.}
  \label{tab:best_comb_cross_val}
\end{table*}

\subsection{Detailed Prompt}
\label{ssec:0s_prompt}

\autoref{fig:prompt_0s} shows the prompt used in our approach. Its formatting depends on the annotated metadata, i.e., whether the instance has annotated answer options and whether the figure consists of multiple subfigures.

\begin{figure*}[t]
    \centering
    \scriptsize
    \fbox{\begin{minipage}{0.97\textwidth}
    \sffamily
    \textbf{System Message:}
        You are an assistant answering questions about (semi-)structured figures such as charts and diagrams. Answer the question as precisely as possible.

    \textbf{User Message:}
        Image: \{image\} \\
        Question: '\{question\}' \\
        \fbox{\parbox{0.97\columnwidth}{
        \texttt{if image\_metadata['answer\_options']:}\\[0.2em]
        \parbox{0.95\columnwidth}{
            \hspace*{1.5em}\parbox{0.9\columnwidth}{
                Answer options: \{answer\_options\}
            }
        }
        }}
        \vspace{0.5em}
        Additional Information: \\
        - The caption of the image is '{image\_metadata['caption']}'.\\

        \fbox{\parbox{0.97\columnwidth}{
        \texttt{if image\_metadata["compound"]:}\\[0.2em]
        \parbox{0.95\columnwidth}{
            \hspace*{1.5em}\parbox{0.9\columnwidth}{
                - The figure image contains \{image\_metadata['figs\_numb']\} (sub)figures which can be separated and constitute individual figures.
            }
        }
        \texttt{else:}\\[0.2em]
        \parbox{0.95\columnwidth}{
            \hspace*{1.5em}\parbox{0.9\columnwidth}{
                - The figure image contains a single figure object which cannot be decomposed into multiple subfigures.
            }
        }
        }}
        
        \vspace{0.5em}
        - The figure type is '\{image\_metadata['figure\_type']\}'.\\
        
        Task: \\ 
        You are presented with a figure and an associated question. \\

        \fbox{\parbox{0.97\columnwidth}{
        \texttt{if image\_metadata['answer\_options:']:}\\[0.2em]
        \parbox{0.95\columnwidth}{
            \hspace*{1.5em}\parbox{0.9\columnwidth}{
                Your task is to select the correct answer options based on the figure. One or more answer options are correct. Only respond with the key(s) of the correct answer option(s), so e.g., 'A,C' if answer options A and C are correct.
            }
        }
        \texttt{else:}\\[0.2em]
        \parbox{0.95\columnwidth}{
            \hspace*{1.5em}\parbox{0.9\columnwidth}{
                Your task is to answer the question based on the figure.
            }
        }
        }}
        
        \vspace{0.5em}
        
        \vspace{0.5em}
        
        You should only use the information in the figure to answer the question. Do not use any external knowledge or information. If the figure does not provide enough information to answer the question, respond with 'It is not possible to answer this question based only on the provided data.'. If you can answer the question, simply provide the answer without further explanation and do not repeat the question.\\
        Answer:        
    \end{minipage}}
    \caption{The zero-shot prompt is formatted based on the annotated metadata via conditional statements. The MLLM is not given the if-else logic; it is only given the indented text inside the block. Bold text is not part of the prompt for the LLM either; it only indicates which parts of the prompt belong to the system or user message. Values in brackets are placeholders for the respective instance's actual values.
    }
    \label{fig:prompt_0s}
\end{figure*}

\subsection{Dataset Characteristics}
\label{ssec:dataset_characteristics}

\begin{figure}[t]
 \centering
 \includegraphics[width=1\columnwidth]{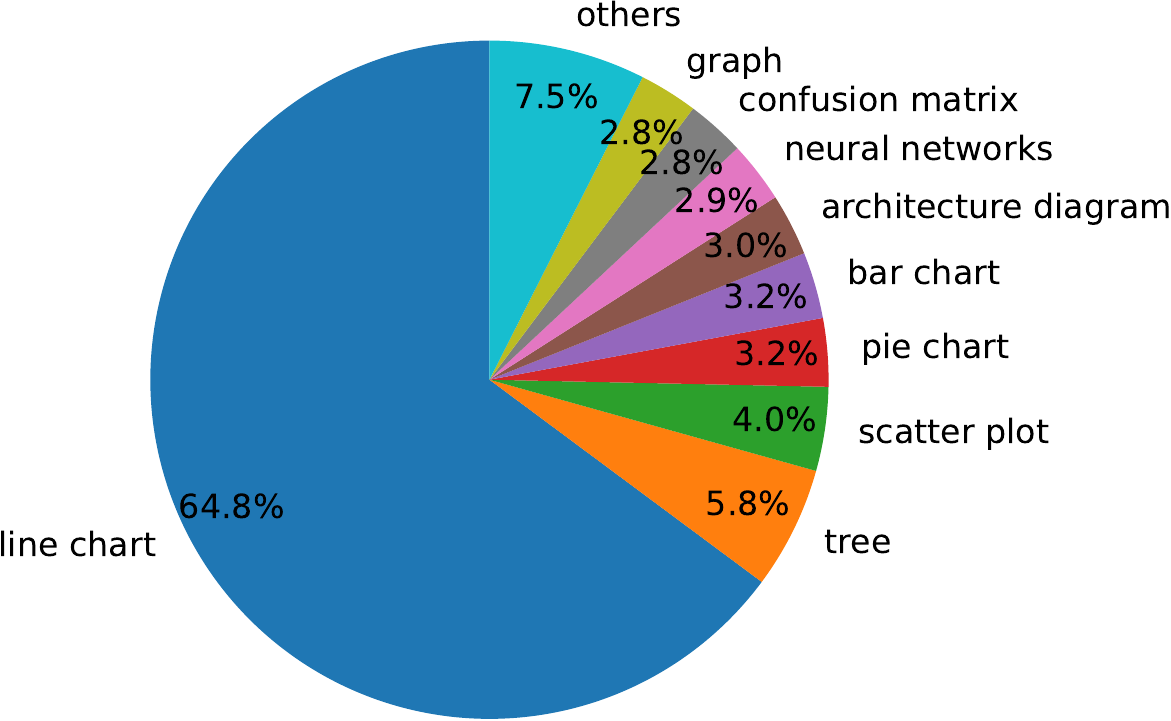}
  \caption {Figure type distribution on development set.} 
  \label{fig:fig_dist_dev}
\end{figure}

The dataset consists of 3000 real-world figures extracted from English scientific publications available in the ACL Anthology and arXiv. The figures can be categorized into different figure types such as \textit{line chart}, \textit{tree}, or \textit{scatter plot}. These figure types are not evenly distributed. For example, \textit{line chart} makes up 65\% of all figures in the development set (see \autoref{fig:fig_dist_dev}). 

Each figure is annotated with seven questions. Two binary questions (one focusing on visual features and one focusing on non-visual features), two questions with four answer options respectively (one visual and one non-visual), two questions with infinite answer sets (one visual and one non-visual), and one unanswerable question. The unanswerable questions are not subdivided into visual and non-visual questions, and they generally follow a different pattern than the answerable ones. For example, they mostly refer to material unavailable to the model and often do not focus on the images' visual/non-visual features.


\end{document}

%% file: commands.tex
\usepackage{csquotes}

\usepackage{todonotes}

\definecolor{anne}{rgb}{0,0.5,0.9}
\definecolor{instr}{rgb}{0.8,0.67,18}
\definecolor{chris}{rgb}{0.5,0.25,0.25}
